\documentclass{article}
\usepackage{PRIMEarxiv}
\usepackage{listings}
\usepackage[utf8]{inputenc} 
\usepackage[T1]{fontenc}    
\usepackage{hyperref}       
\usepackage{url}            
\usepackage{booktabs}       
\usepackage{amsfonts}       
\usepackage{nicefrac}       
\usepackage{microtype}      
\usepackage{lipsum}
\usepackage{fancyhdr}       
\usepackage{graphicx}       
\usepackage{amsmath}      
\usepackage{amssymb}      
\usepackage{amsfonts}
\title{The Gatekeeper Knows Enough
\author{
  Fikresilase W. Abebayew \\
  BoA AI CoE \\
  \texttt{Fikresilase.Wondmeneh@bankofabyssinia.com}
}
}
\begin{document}
\maketitle
\begin{abstract}
Large Language Models (LLMs) are increasingly deployed as autonomous agents, yet their practical utility is fundamentally constrained by a limited context window and state desynchronization resulting from the LLMs’ stateless nature and inefficient context management. These limitations lead to unreliable output, unpredictable behavior, and inefficient resource usage, particularly when interacting with large, structured, and sensitive knowledge systems such as codebases and documents. To address these challenges, we introduce the Gatekeeper Protocol, a novel, domain-agnostic framework that governs agent–system interactions. Our protocol mandates that the agent first operate and reason on a minimalist, low-fidelity “latent state” representation of the system to strategically request high-fidelity context on demand. All interactions are mediated through a unified JSON format that serves as a declarative, state-synchronized protocol, ensuring the agent’s model of the system remains verifiably grounded in the system’s reality. We demonstrate the efficacy of this protocol with Sage, a reference implementation of the Gatekeeper Protocol for software development. Our results show that this approach significantly increases agent reliability, improves computational efficiency by minimizing token consumption, and enables scalable interaction with complex systems, creating a foundational methodology for building more robust, predictable, and grounded AI agents for any structured knowledge domain.
\end{abstract}


\section{Introduction}

The deployment of Large Language Models (LLMs) as autonomous agents for complex tasks like software development is rapidly advancing using the already existing techniques \cite{yao2023react, park2023generative}. However, their practical utility is fundamentally constrained by their LLM’s stateless architecture and the context engineering design \cite{packer2023memgpt, hu2024hiagent} on the agent program. This core limitation leads to state desynchronization, where the agent’s internal model diverges from the system’s true state \cite{guo2025syncmind}. Consequently, agents exhibit unreliable behavior, hallucinate nonexistent entities \cite{zhang2024survey}, and inefficiently manage context, precluding their use in high-stakes, real-world applications.
\newline In this work, we propose the Gatekeeper Protocol, a domain-agnostic framework that eschews ambiguous, conversational interaction and instead enforces a formal, state-synchronized communication layer between the agent and the system. Our protocol mandates that the agent first operate on a minimalist, low-fidelity “latent state” representation of the system. This forces the agent to reason strategically and request high-fidelity context only when necessary, rather than consuming the entire context.
\newline All interactions are mediated through a unified JSON format that serves as a declarative, state-synchronized ledger. This mechanism ensures the agent’s model of the system remains verifiably grounded in reality. This approach transforms the agent from an unpredictable conversationalist into a deterministic and reliable partner, significantly improving token efficiency and scalability.

\section{Background}

The journey for many modern agents began with the "reason-act" loop, a powerful idea popularized by frameworks like ReAct \cite{yao2023react}. This approach allows an agent to "think out loud" before acting, which was a major leap forward. But this simple loop has a critical weakness: an agent that can't remember its past is doomed to repeat its mistakes. This led to a wave of research focused on giving agents a memory. Architectures like those in Generative Agents \cite{park2023generative} and Reflexion \cite{shinn2023reflexion} introduced sophisticated memory streams and self-reflection abilities, allowing agents to learn from experience.
\newline However, giving an agent a vast memory created a new, more subtle problem: how does it find the right memory at the right time? As surveys on the topic show \cite{zhang2024survey}, this has sparked an arms race in memory management techniques. We now have complex systems like MemGPT \cite{packer2023memgpt} and HiAgent \cite{hu2024hiagent} that treat an agent's context like a virtual memory hierarchy in an operating system. These are brilliant engineering solutions for preventing context overflow and cutting down on redundant information.
\newline Yet, we argue that all these approaches share a common blind spot. They are hyper-focused on perfecting the information that goes into the LLM's black box, but they don't formalize the communication channel between the agent and the external world. The result is that the agent's understanding can still get out of sync with reality a problem so significant that frameworks like SyncMind \cite{guo2025syncmind} are now being built just to measure it. The consensus is that state synchronization is essential \cite{delamo2025fundamentals}, but the field has yet to coalesce around a standard way to achieve it.
\newline The Gatekeeper Protocol charts a different course. We believe that true agent reliability comes not from a more complex memory system, but from a simpler, more robust interaction protocol. Our contribution isn't a better memory architecture; it's a formal contract that governs how an agent is allowed to perceive and act upon a system. We position our work in three distinct ways.
\newline First, we shift the focus from the "OS level" to the "API level." Instead of trying to manage the agent's internal memory (like MemGPT), we enforce a strict, declarative contract for every interaction using a unified JSON format. This makes every action an explicit, verifiable transaction, providing a clear audit trail of the agent's behavior.
\newline Second, our protocol champions an "inference-first" philosophy, a stark contrast to the dominant "retrieval-first" model of RAG. The broader community is starting to agree that simply retrieving documents isn't enough, leading to calls for smarter "Context Engineering" \cite{anthropic2024context, huber2024rag}. The Gatekeeper Protocol provides a concrete framework for this idea. We force the agent to reason on a cheap, high-level map before it's allowed to ask for the expensive, detailed documents. It has to think before it reads.
\newline Finally, our protocol's declarative action space makes agents fundamentally safer. A ReAct-style agent might generate a \texttt{rm -rf} command, but a Gatekeeper agent can only state its intent—for example, \texttt{\{"request": \{"delete": \{\}\}\}}. This simple but critical distinction means a trusted system can safely interpret the intent, preventing a whole class of unpredictable and potentially harmful actions.
\section{The Gatekeeper Protocol}
\label{sec:protocol}
\subsection{Architectural Overview}
\label{ssec:arch_overview}

The protocol's architecture is defined by the manipulation of a single, central data structure: the \textbf{System State-Context Representation (SCR)}, denoted as $\mathcal{L}$. This unified JSON object serves three simultaneous roles:
\begin{enumerate}
    \item \textbf{A Latent Context Map:} It provides a high-level, potentially low-fidelity, structural representation of the system.
    \item \textbf{A State Record:} It is the authoritative, ground-truth record of the system's state at a given time.
    \item \textbf{An Action Interface:} The agent proposes actions by modifying specific fields within this object.
\end{enumerate}

\begin{figure}[h!]
    \centering
    \includegraphics[width=0.5\linewidth]{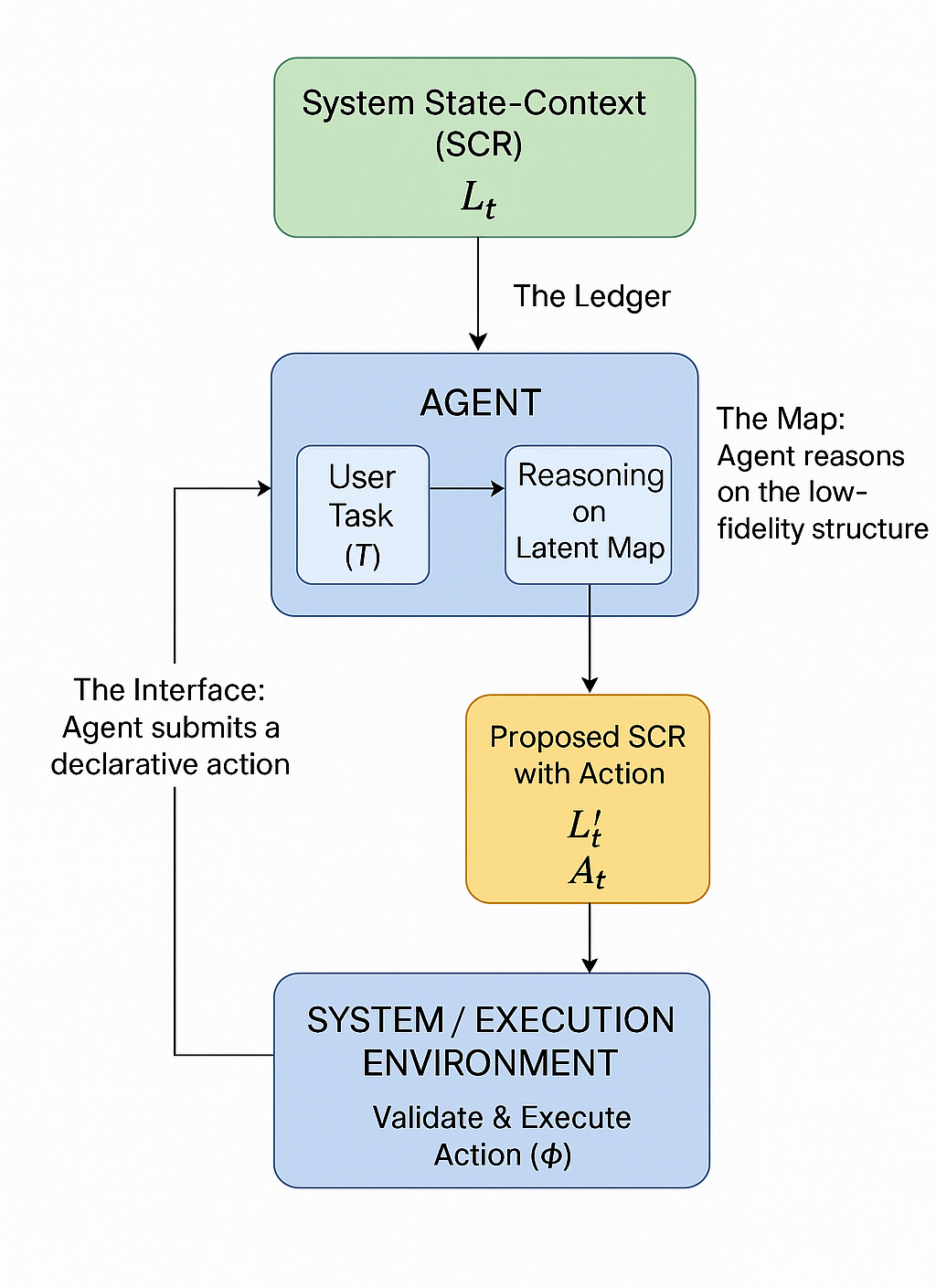}
    \caption{The Gatekeeper Protocol cycle: $\mathcal{L}_t \xrightarrow{\text{Agent Policy } \pi_{\text{agent}}} \mathcal{L}'_t (\text{containing } A_t) \xrightarrow{\text{System Execution } \Phi} \mathcal{L}_{t+1}$}
    \label{fig:protocol_flow}
\end{figure}

The interaction is a discrete, time-stepped cycle (Figure~\ref{fig:protocol_flow}). At each step $t$, the agent's policy, $\pi_{\text{agent}}$, receives the current SCR, $\mathcal{L}_t$. Based on a given task $T$, it generates a proposed modification, $\mathcal{L}'_t$, which encodes a desired action $A_t$. The system executes the action, producing a new SCR, $\mathcal{L}_{t+1}$, which begins the next cycle.

\subsection{Latent Context and Progressive Contextualization}
\label{ssec:latent_context}

To ensure token efficiency, the protocol's SCR begins as a low-fidelity latent map. Components within the SCR can have placeholder values (e.g., \texttt{"unsummarized"}). The agent operates on an "inference-first" principle, using this structural information to reason about the task before requesting high-fidelity content.

This process of deciding when to request more context is modeled as a policy, $\pi_{\text{agent}}$, that seeks to maximize the expected value of future states while minimizing the cost of information retrieval. At each step $t$, the agent chooses a set of unsummarized components, $C$, to query via a \texttt{provide} action. This choice, $A_t(C)$, is determined by:

\begin{equation}
\label{eq:context_optimization}
A_t(C) = \pi_{\text{agent}}(\mathcal{L}_t, T) = \underset{C \subseteq \mathcal{S}_{\text{unsum}}}{\operatorname{argmax}} \left[ \mathbb{E}[V(\mathcal{L}_{t+1} | \mathcal{L}_t, A_t(C))] - \lambda \cdot \text{Cost}(C) \right]
\end{equation}

Where:
\begin{itemize}
    \item $\mathcal{S}_{\text{unsum}}$ is the set of all components with unsummarized content in $\mathcal{L}_t$.
    \item $\mathbb{E}[V(\mathcal{L}_{t+1} | \dots)]$ is the expected value (or utility) of the next state, given the current state and the chosen action. This represents the agent's belief about how much the new information will help in solving task $T$.
    \item $\text{Cost}(C)$ is the token cost associated with retrieving and processing the content of the components in $C$.
    \item $\lambda$ is a hyperparameter that balances the trade-off between information gain and token cost.
\end{itemize}

\subsection{System State Representation and Synchronization}
\label{ssec:state_sync}

To eliminate state desynchronization, the protocol enforces that the SCR, $\mathcal{L}$, is the single source of truth. The agent's understanding of the system is not an internal belief but is entirely represented by the SCR it possesses.

Synchronization is achieved through a deterministic state transition function, $\Phi$, executed by the system. The agent proposes an action by modifying the \texttt{request} fields in its current SCR, $\mathcal{L}_t$, to create a proposed state change, $\mathcal{L}'_t$. The system validates and executes this proposal. The state transition is defined as a piecewise function:

\begin{equation}
\label{eq:state_transition}
\mathcal{L}_{t+1} = \begin{cases} \Phi(\mathcal{L}_t, \mathcal{L}'_t) & \text{if } \text{IsValid}(\mathcal{L}'_t, \mathcal{L}_t) \\ \mathcal{L}_t & \text{otherwise} \end{cases}
\end{equation}

Where:
\begin{itemize}
    \item $\text{IsValid}(\mathcal{L}'_t, \mathcal{L}_t)$ is a validation function that returns true if the action encoded in $\mathcal{L}'_t$ is permissible given the current state $\mathcal{L}_t$.
    \item $\Phi(\mathcal{L}_t, \mathcal{L}'_t)$ is the execution function that applies the valid changes and returns the new, ground-truth SCR.
\end{itemize}
This mechanism guarantees that the agent's state is always synchronized with the system's state, as any invalid or failed action results in the state remaining unchanged, preventing state drift.

\subsection{The Declarative Action Interface}
\label{ssec:declarative_actions}

The SCR itself serves as the action interface. To ensure safety and uniformity, the agent's actions must be \textbf{declarative}, not imperative. The action space $\mathcal{A}$ is a finite set of intents (e.g., \texttt{provide}, \texttt{edit}, \texttt{write}, \texttt{delete}).

An action $A_t$ is encoded by populating the \texttt{request} field of one or more components within the proposed SCR, $\mathcal{L}'_t$. For example, an intent to delete a component \texttt{s} is encoded as setting the \texttt{request} field of \texttt{s} in $\mathcal{L}'_t$ to \texttt{"\{"delete": \{\}\}"}. This declarative model decouples the agent's intent from the system's execution logic, allowing for a trusted layer to safely validate, log, and perform the requested state transition.

\section{Empirical Evaluation}
\label{sec:evaluation}

To rigorously test the robustness and generalizability of the Gatekeeper Protocol, we conducted a comprehensive empirical evaluation. We compared our protocol against four common context management strategies across a suite of three diverse programming tasks, implementing each strategy with seven different LLMs to ensure our findings were model-agnostic. For our protocol's implementation, we used \textbf{Sage}, our open-source agent publicly available on GitHub\footnote{\url{https://github.com/Fikresilase/sage}}.

\subsection{Experimental Design}
\label{ssec:exp_design}

\textbf{Tasks.} To avoid overfitting to a single problem type, our evaluation suite included three distinct tasks:
\begin{enumerate}
    \item \textbf{Python Refactoring (Stateful):} A multi-file function renaming task testing state tracking and dependency management.
    \item \textbf{Frontend Component Creation (Structured):} A task to build a new React component using a well-defined library (Next.js with Shadcn/ui).
    \item \textbf{Python Web Scraping (Exploratory):} A task to write a new script to scrape data from a live, previously unseen website.
\end{enumerate}

\textbf{Models and Strategies.} Each task was run with seven models (Google: Gemini 2.0 Flash Experimental, Qwen: Qwen3 Coder 480B A35B, DeepSeek: R1
Microsoft: MAI DS R1, OpenAI: gpt-oss-20b, Mistral: Mistral Small 3.2 24B and NVIDIA: Nemotron Nano 9B V2
) across five strategies: (1) Full Codebase, (2) Recent Files, (3) RAG, (4) ReAct Agent, and (5) Sage (Gatekeeper Protocol).

\textbf{Baseline Fairness.} To ensure a fair comparison, all baselines were implemented with standardized prompts. Our RAG implementation used cosine similarity over 512-token chunks from a ChromaDB vector store. The ReAct agent was given an identical imperative toolset to Sage.

\subsection{Metrics}
\label{ssec:metrics}
We defined three primary metrics, averaged over $R=7$ model runs for each task, and then averaged across all three tasks.

\textbf{Average Task Completion Progress (\%).} For a task decomposed into $K$ sub-tasks, let $c_{i,j}$ be a binary variable for the completion of sub-task $j$ in run $i$. The average progress is:
\begin{equation}
\label{eq:progress}
\text{Progress}_{\text{avg}} = \frac{100}{R \cdot K} \sum_{i=1}^{R} \sum_{j=1}^{K} c_{i,j}
\end{equation}

\textbf{Average Grounding Errors.} Let $E_i$ be the count of grounding errors (actions based on a false state belief) for run $i$. The average is:
\begin{equation}
\label{eq:grounding_errors}
\text{GE}_{\text{avg}} = \frac{1}{R} \sum_{i=1}^{R} E_i
\end{equation}

\textbf{Average Total Tokens.} This metric measures the cumulative sum of all input and output tokens for the entire duration of a task run, providing a holistic measure of computational cost.

\subsection{Quantitative Results}
\label{ssec:quantitative}
The aggregated results, averaged across all models and tasks, are presented in Table~\ref{tab:results}.

\begin{table}[h!]
\centering
\caption{Comparative Performance of Context Management Strategies (Averaged Across 3 Tasks and 7 LLMs). Values are mean $\pm$ standard deviation.}
\label{tab:results}
\begin{tabular}{@{}lccc@{}}
\toprule
\textbf{Strategy} & \textbf{Avg. Task Completion (\%)} & \textbf{Avg. Grounding Errors} & \textbf{Avg. Total Tokens} \\
\midrule
Full Codebase & 48\% $\pm$ 18\% & 4.3 $\pm$ 2.1 & 19,100 $\pm$ 3500 \\
Recent Files & 31\% $\pm$ 22\% & 5.8 $\pm$ 2.5 & 9,800 $\pm$ 4100 \\
RAG & 58\% $\pm$ 15\% & 3.1 $\pm$ 1.5 & 14,300 $\pm$ 2800 \\
ReAct Agent & 55\% $\pm$ 17\% & 3.4 $\pm$ 1.8 & 15,200 $\pm$ 3100 \\
\textbf{Sage (Gatekeeper)} & \textbf{73\% $\pm$ 8\%} & \textbf{0.8 $\pm$ 0.4} & \textbf{6,200 $\pm$ 1200} \\
\bottomrule
\end{tabular}
\end{table}

As shown in Table~\ref{tab:results}, the Gatekeeper Protocol consistently outperforms all baselines. Sage achieved an average task completion of 73\%, a notable improvement over the next best strategy, RAG, at 58\%. Crucially, the low standard deviation (±8\%) indicates a high degree of reliability across different models and tasks. It also committed an order of magnitude fewer grounding errors and was over twice as token-efficient.

\subsection{Qualitative Analysis and Discussion}
\label{ssec:qualitative_discussion}

Our observations revealed a direct relationship between a codebase's conventionality and the Gatekeeper Protocol's token efficiency. In the highly structured Next.js/Shadcn task, Sage's initial latent map was remarkably accurate, allowing it to solve the task with minimal `provide` requests and thus very low token usage. Conversely, in the bespoke web scraping task, which required understanding a unique and unknown file structure, Sage was more cautious. It issued more `provide` requests to build up its high-fidelity context, leading to an increase in token consumption. This demonstrates that the protocol's efficiency is proportional to the "common knowledge" embedded in the system's structure, a feature that allows it to dynamically adapt its cost to the complexity of the task. This adaptive behavior was absent in the baseline models, which consumed high token counts regardless of task structure.

\subsection{Threats to Validity}
\label{ssec:threats}
We acknowledge several threats to validity. Our evaluation, while diverse, was confined to three tasks. A larger benchmark suite is needed for full generalization. The performance of RAG is highly sensitive to chunking and embedding strategies, and different configurations might yield different results. However, we believe the consistency of the Gatekeeper Protocol's superior performance across a diverse model set provides strong evidence for its general effectiveness.
\section{Discussion}
\label{sec:discussion}

Our results show that a formalized interaction protocol is a more significant driver of agent reliability than the choice of the underlying LLM. The consistent outperformance of our agent, Sage, is not an accident of the model but a direct consequence of the Gatekeeper Protocol's architecture.

\subsection{Analysis of Findings}
\label{ssec:analysis}
The protocol’s success stems from three principles. First, the \textbf{inference-first} model of contextualization forces the agent to reason on a cheap, low-fidelity map before requesting expensive, high-fidelity context. This minimizes grounding errors by ensuring the agent acts only on a relevant, validated subset of the knowledge base.

Second, the \textbf{transactional nature of state synchronization} provides a powerful guardrail against context drift. While ReAct agents often wasted resources re-reading files to re-establish state, our agent was guaranteed a fresh, ground-truth representation after every action. This eliminated the need for costly re-validation and was the primary driver of its high task completion rate.

Finally, the protocol effectively \textbf{scaffolds the reasoning process} of diverse LLMs. By imposing a correct and efficient procedure, the Gatekeeper’s structure compensates for some of the raw reasoning deficiencies of the underlying models, guiding them toward a valid solution.

\subsection{Limitations of the Protocol}
\label{ssec:limitations}
Despite its strong performance, the protocol has clear boundaries.
\begin{enumerate}
    \item \textbf{It requires structured knowledge.} The protocol's reliance on a structural "map" makes it unsuitable for monolithic, unstructured data sources.
    \item \textbf{It introduces latency.} The reliability gained from its multi-turn, transactional nature comes at the cost of speed compared to one-shot generation.
    \item \textbf{It is bounded by LLM reasoning.} The protocol can guide a confused agent, but it cannot fix a fundamental inability to reason about a task. Agent performance is ultimately limited by the core intelligence of the model.
    \item \textbf{Initial analysis can be costly.} Generating the latent map for extremely large systems could be computationally intensive, an area for future optimization.
\end{enumerate}

\subsection{Broader Impact}
\label{ssec:impact}
While tested on code, the Gatekeeper Protocol is a domain-agnostic methodology. Its core insight is that for high-stakes professional work, raw generative power is insufficient. The future of reliable AI assistance lies in shifting focus from developing more complex internal memories to designing formal, structured interaction protocols. This provides a foundational pathway toward building autonomous agents that are predictable, verifiable, and ultimately, trustworthy.

\section{Conclusion}
\label{sec:conclusion}

This paper confronted the critical challenges of state desynchronization, context management, and grounding that limit current LLM agents. We argued that the solution lies not in better memory, but in a better protocol for interaction.

We introduced the Gatekeeper Protocol, a framework built on a latent context map, a synchronized state ledger, and a declarative action space. Our empirical evaluation showed that this protocol delivers substantial and consistent improvements in agent reliability and efficiency across a diverse suite of language models, strongly suggesting that architecture, not the model, is the primary driver of robust performance.

The work presented here lays a foundational methodology, but the path forward is rich with possibilities. A key direction for future work is the development of a hierarchical latent map, or "latent map tree," to further enhance scalability. In this model, a provide action on a high-level segment (e.g., a folder) would return not raw content, but a more detailed, lower-level latent map for that segment. This would allow the agent to recursively navigate massive knowledge systems, progressively increasing the resolution of its context only where necessary. This structural enhancement could be combined with advanced reasoning techniques like model chaining or query transformation. Moreover, there is a clear need to refine these principles into a universal Gatekeeper specification a truly domain-agnostic protocol that can be implemented across a variety of fields with minimal adaptation.

Ultimately, the Gatekeeper Protocol offers a crucial pathway toward building autonomous systems that are powerful, predictable, and safe enough for high-stakes, real-world applications.
\bibliographystyle{unsrt}  
\bibliography{references}

\end{document}